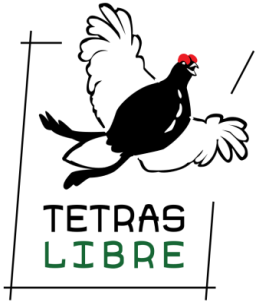
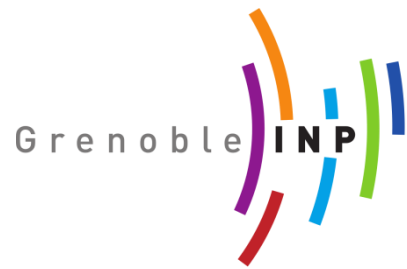
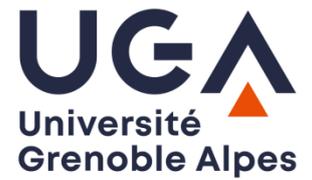

# Research report for the UNseL project

## Transforming UNL graphs in OWL representations

*Transformations de graphes UNL en représentations OWL*

| | |
|---|---|
| Version: | 1 |
| Last edited: | 09/01/2022 |
| Author: | David.Rouquet@tetras-libre.fr |
| Co-authors: | Valérie Bellynck, Vincent Berment and Christian Boitet |

# Sommaire

**Abstract.** Extracting formal knowledge (ontologies) from natural language is a challenge that can benefit from a (semi-) formal linguistic representation of texts, at the semantic level. We propose to achieve such a representation by implementing the Universal Networking Language (UNL) specifications on top of RDF. Thus, the meaning of a statement in any language will be soundly expressed as a RDF-UNL graph that constitutes a middle ground between natural language and formal knowledge. In particular, we show that RDF-UNL graphs can support content extraction using generic SHACL rules and that reasoning on the extracted facts allows detecting incoherence in the original texts. This approach is experimented in the UNseL project that aims at extracting ontological representations from system requirements/specifications in order to check that they are consistent, complete and unambiguous. Our RDF-UNL implementation and all code for the working examples of this paper are publicly available under the CeCILL-B license at https://gitlab.tetras-libre.fr/unl/rdf-unl

# Contexte

La société Tétras Libre – SARL réalise des prestations pour le compte de l'Institut polytechnique de Grenoble (Grenoble INP) dans le cadre du projet RAPID UNseL.

Ce document fait partie des rendus de Tétras Libre à Grenoble-INP pour le Lot 1 : *Études amont et préliminaires*. Il traite de l'étude des transformations entre graphes UNL et représentations logiques formelles (OWL) et propose une version Web Sémantique d'UNL basée sur RDF. Le rapport est rédigé en anglais pour servir de base à de futurs articles de recherche.

# Licence et propriété

Ce document est propriété de Grenoble-INP.

Il est distribué le 15/05/2020 aux partenaires du projet UNseL sous licence CeCILL-B (https://cecill.info/licences/Licence_CeCILL-B_V1-fr.html).

Cette licence est compatible avec la licence BSD et autorise toute exploitation de ce document, même pour dériver des contenus sous licence fermée. Elle oblige toutefois à citer de façon claire les auteurs et propriétaires dans les crédits des réutilisations, soit :

**Rapport de recherche du projet UNsel.**
**Auteur :** David Rouquet pour Tétras Libre – SARL (https://www.tetras-libre.fr/)
**Propriétaire :** Institut polytechnique de Grenoble (Grenoble INP)

**Auteur : David Rouquet pour Tétras Libre – SARL**

**Raison sociale :** Tétras libre – SARL
**Adresse :** 8 rue de Mayencin, 38 400, Saint Martin d'Hères, France
**Siège :** 464 Route d'Uriage, 38 410, Saint Martin d'Uriage, France
**N° SIRET :** 825 047 541 00024

**Site Web :** https://www.tetras-libre.fr/

# 1   Introduction and motivations

The UNseL[1] project aims at providing tools to automatically detect ambiguity, incoherence and incompleteness in system requirements and specifications, written in natural languages. We typically target the design of telecommunication systems or aeronautical equipment but are not limited to those. To check the quality of requirements and specifications, especially their coherence, we must extract and represent the meaning of texts in a form that can support automatic reasoning. This task is directly related to the broader Artificial Intelligence problem of *machine reading*, the automatic understanding of texts, defined in [1] as : "*the formation of a coherent set of beliefs based on a textual corpus and a background theory*".

Among other tools and formalisms that shall be used to support the reasoning part of the project, the Semantic Web standard stack has been identified as the most promising. It provides: (1) convenient formats and vocabularies for interoperability at the syntactic and semantic levels (RDF, RDFS, OWL, etc.), (2) generic reasoning tools to check consistency and infer facts in knowledge bases and (3) rule languages to develop custom transformations and validations (SHACL and SPARQL).

This paper focuses on bringing to the Semantic Web a linguistic framework to soundly represent the meaning of a sentence, directly in RDF. It proposes a complete serialization of the *Universal Networking Language* (*UNL*) as a schema on top of RDF (we simply call it RDF-UNL). UNL is a linguistico-semantic interlingua that represents a sentence in any natural language L as a hypergraph, where arcs bear semantic relations, and nodes bear interlingual lexemes (word senses) taken from an autonomous lexical space, plus semantic and pragmatic features. It has been acknowledged as a framework suited to machine translation and tasks such as (multilingual) question answering, information extraction, information retrieval, etc. [2] Therefore it is a strong candidate to operate as a linguistic paradigm for machine reading in Semantic Web applications.

On the other hand, our proposal will allow Natural Language Processing tools based on UNL to natively benefit from Semantic Web technologies and resources. It will create a mutually beneficial bridge between those research areas.

The paper is organized as follows. In Section 2, we present UNL and give details about its components and their implementation on top of RDF. In section 3, we show how RDF-UNL can be used to represent the meaning of sentences and documents and discuss the implementation of an advanced feature of UNL, the *scopes*. Finally, section 0 presents a working example that illustrates how to extract OWL statements from a RDF-UNL representation of two system requirements. An automatic reasoning is performed on the OWL statements to exhibit incoherence in the requirements. This last example, even if preliminary, shows the potential of our proposal to complement existing methods [3–5] in extracting meaningful knowledge from texts.

---

[1]   UNseL : Universal Networking system engineering Language

Our RDF-UNL implementation and all the code for the working example are publicly available under the CeCILL-B license[2] in a Gitlab repository at [6]. All code samples in this article are using the RDF Turtle syntax [7], except if stated otherwise.

## 2 UNL components and implementation in RDFS

### 2.1 Generalities about UNL

The UNL program started in 1996, as an initiative of the *Institute of Advanced Studies*[3] of the United Nations University in Tokyo, Japan. Its goal is to construct a multilingual infrastructure so all peoples can have access to information and knowledge in their native language and culture.

The UNL language represents the meaning of a sentence as a hyper-graph, as we illustrate in Fig. 1, where:

- The nodes are so-called *Universal Words* (UW) completed with semantic and pragmatic attributes (time, modality, deep aspect, etc.) or sub-graphs (*scopes*) representing parts of the sentence,

- The directed arcs are labeled with semantic binary relations (agent, beneficiary, duration, destination, etc.), linking together nodes and/or sub-graphs (*scopes*).

Because UNL structures use only binary relations and unary attributes, they can be represented very naturally with RDF triples.

---

[2]   https://cecill.info/licences.fr.html

[3]   https://ias.unu.edu/en/

*R1: The system allows a channel to take on two states: listening and traffic*

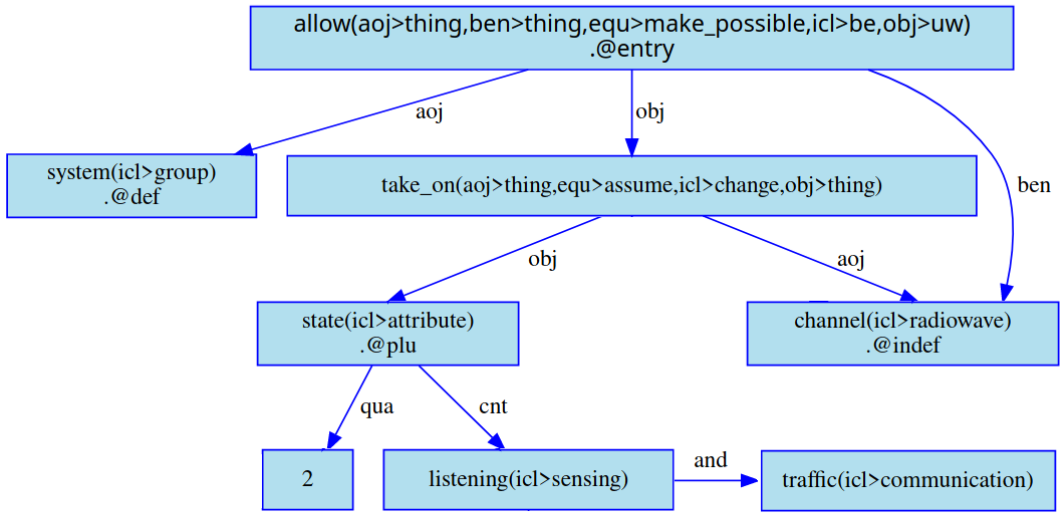

*Fig. 1. Example of simple UNL graph*

The UNL specifications have evolved over time, mainly to expand or reconsider the sets and definitions of allowed relations and attributes in UNL graphs. The different versions are available in [8]. Our current work focused on the last version (UNL2010) proposed by the UNL[Web] community [9]. We plan to also release at least version 3.3 available as an aligned RDF scheme so our work can apply to systems that are not aligned with the UNL2010 proposal.

The following sections present in detail the fundamental components of the UNL and their implementation as parts of an RDF scheme. The implementation should be self-sufficient, containing all necessary definitions, examples and external links. The scheme makes use of the standard vocabularies RDFS [10], SKOS [11] and OWL [12]. The namespace of the scheme is: @prefix unl: <https://unl.tetras-libre.fr/rdf/schema#> .

We used two environments for the development : the Open Source OWL editor Protégé 5[4] [13] and the commercial Topbraid Composer that offers a free version[5].

## 2.2   Universal Words and the UNL Knowledge Base (UNLKB)

The vocabulary of UNL is composed of interlingual lexemes called *Universal Words* (*UW* for short). A UW is designed to refer unambiguously to a concept, shared among several languages. However, UWs correspond to acceptions (word senses) in a language, so distinct UWs can exist for *affection* and *disease*, even though they refer to the same concept. A UW is made of:

---

[4]  https://protege.stanford.edu/

[5]  https://www.topquadrant.com/topbraid-composer-install/

1. A *headword*, if possible derived from English, which can be a word, initials, a compound word, an expression or even an entire collocation. It is a label for the concepts it represents in its original language,
2. A *list of restrictions* that aims to precisely specify the concept the UW refers to. A restriction is composed of a word linked by a Universal Relation (see section 2.3)

The following are possible UWs:

- play(icl>act,agt>thing,obj>thing) and play(icl>show)

(the sense of the headword is focused by the attributes, *icl* stands for *included in*).

- ikebana(icl>flower_arrangement) -- (the headword comes from Japanese).
- *go_down* -- (the headword does not need any refinement)

For a complete volume of UWs, any word appearing in a UW restriction should be the headword of another UW (except for top concepts defined below). This leads to the following notion: The *Master Definition* of a UW expands the restrictions with complete UWs instead or only headwords. For instance, the master definition of *play(icl>show)* is *play{icl>show(icl>thing)}*. The UW can also be noted with the abbreviation *play(icl>show>thing)*.

We see that master definitions organize naturally a volume of UWs in a semantic network called a *UNL Knowledge Base* (*UNLKB*). A UNLKB can be qualified as a pre-ontology, as it arranges word senses instead of concepts and its logical validity is not ensured. Relations *icl* (included in), *iof* (instance of) and *equ* (equivalent) play a special role in structuring a KB and can be related resp. to *rdfs:subClassOf*, *rdfs:instanceOf* and *owl:sameAs*. UWs can be anchored to top level concepts in the KB that are not associated with UWs.

**Fig.** 2 illustrates a set of possible top concepts :

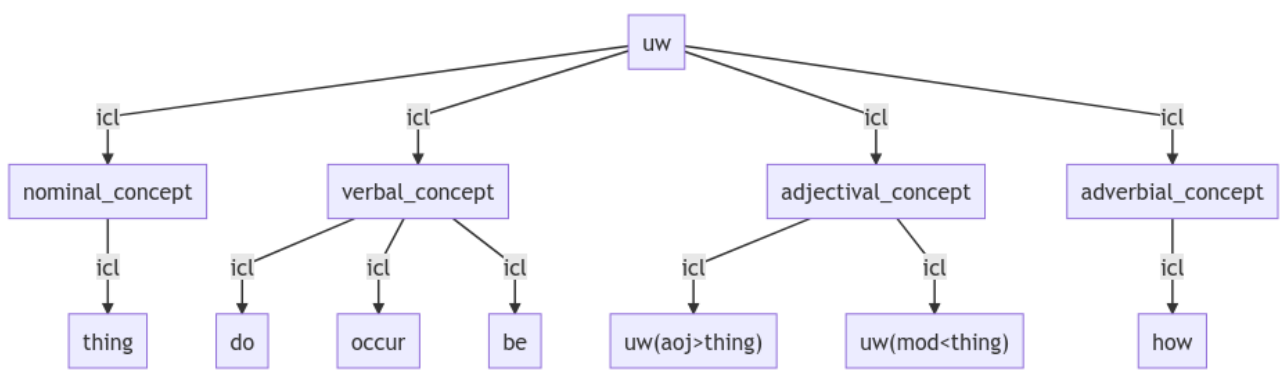

*Fig. 2. Example of top concepts in a UNLKB*

Several volumes of UWs have been created and linked to various natural languages. For instance, the following are freely available on the Web:

- A volume initiated from Princeton Wordnet 2.1 synsets and post edited[6],
- A volume by the UNL^Web community, available after creating an account[7].

**Implementation.** In the RDF-UNL schema, we created a class *unl:Universal_Word* with the following sub-classes :

- *unl:UW_Lexeme*, for the UWs themselves, it must be divided in subclasses for different volumes of UWs,
- unl:UW_Occurrence, for instance, in the text "the black cat and the white cat", two occurrences cat:01(icl>mammal) and cat:02(icl>mammal) will appear in the UNL graph for the same UW cat(icl>mammal),
- *unl:UNLKB_Top_Concept*, even though they are not real UWs, it was handy to subsume them here.

The following properties apply:

- In a RDF-UNL graph representing a sentence, *unl:UW_Occurrence* instances are linked with *Universal Relations* (see section 2.3) and support *Universal Attributes* (see section 2.3).
- In a UNLKB, *unl:UW_Lexeme* and/or *unl:UNLKB_Top_Concept* instances are linked with *unl:Universal_Relations* (see section 2.3).
- Instances of unl:UW_Lexeme have an owl:AnnotationProperty named unl:has_id and unl:has_master_definition. The UW lexical form is stored using rdfs:label.
- Corresponding *unl:UW_Lexemes* and *unl:UW_Occurrences* appearing respectively a UW volume and a RDF-UNL graph are linked with the following pieces of *owl:ObjectProperty*:
    - unl:has_occurrence and
    - unl:is_occurrence_of (the two are linked by owl:inverseOf).

---

[6] https://gitlab.com/dikonov/Universal-Dictionary-of-Concepts

[7] http://www.unlweb.net/unlarium/

The following presents an extract of code related to a *unl:UW_Lexeme* instance. For readability, we chose to derive the URI from the lexeme and not from the id:

```
@prefix example: <https://unl.tetras-libre.fr/rdf/example#> .
example:broadcast(icl--message)
    a unl:UW_Lexeme , example:Test_UW_Volume ;
    rdfs:label "broadcast(icl>message)" ;
    unl:has_master_definition "broadcast{icl>message(icl>thing)}" ;
    unl:has_id "202004223698" ;
    unl:has_occurrence example:broadcast(icl--message)__00000016 ;
    unl:icl <https://unl.tetras-libre.fr/rdf/example#message(icl--thing)> .
```

Further examples are shown in section 3, presenting how instances of *unl:UW_Occurrence* are used to represent sentences as RDF-UNL graphs.

## 2.3 Universal relations

Universal relations describe semantic binary and directed relations between two nodes of a UNL graph. These relations are at the conceptual level and are more abstract than grammatical relations, even though they can strongly overlap in some cases, like the semantic agent and the grammatical subject (but only when it is a volitional agent).

UNL2010 specifications propose 40 universal relations organized in a hierarchy. User can refer to the specifications at [14] or to RDF-UNL at [6] for the complete list.

**Implementation.** We begin by defining the class *unl:UNL_Node*, a concept that does not exist in UNL specifications but that is handy for us, especially to declare relations domains and ranges. *unl:UNL_Node* contains of the following subclasses :

- *unl:UNLKB_Node* (nodes appearing in a UNLKB)
    - *unl:UW_Lexeme*
    - *unl:UNLKB_Top_Concept*
- *unl:UNL_Graph_Node* (nodes appearing in a graph representing a sentence)
    - *unl:UW_Occurrence*
    - *unl:UNL_Scope*

Universal relations are at first declared as instances of *owl:ObjectProperty* structured as a hierarchy with *rdfs:subPropertyOf*. The root of this hierarchy is the generic *unl:Universal_Relation*. It also made sense to declare this generic root property as a sub-property of *skos:semanticRelation*. Relations can be used in two contexts:

- A UNLKB, as we saw in section 2.2, linking instances of *unl:UNLKB_Node*,
- A RDF-UNL graph representing a sentence, as we will see in section 3, linking instances of *unl:UNL_Graph_Node*.

The following presents an extract of code related to the definition of a *unl:Universal_Relation* :

```
unl:ant    a owl:Class ;
           a owl:ObjectProperty ;
           rdfs:label "ant" ;
           rdfs:subPropertyOf :Universal_Relation , unl:aoj ;
           rdfs:domain unl:UNL_Node ;
           rdfs:range  unl:UNL_Node ;
           skos:altLabel "opposition or concession"@en ;
           skos:definition "   Used to indicate that two entities do not share the same
                    meaning or reference. Also used to indicate concession."@en ;
           skos:example """John is not Peter = ant(Peter;John) / 3 + 2 != 6 = ant(6;3+2) /
                    Although he's quiet, he's not shy = ant(he's not shy;he's quiet)"""@en .
```

Further examples are shown in section 3, presenting how sub-properties of *unl:Universal_Relation* are used to represent sentences as RDF-UNL graphs. We will see that the usage of *scopes* makes things slightly more complicated and will propose adjustments in the representation of universal relations to take this into account.

## 2.4   Universal attributes

Attributes are unary properties on *unl:UNL_Graph_Node* instances. They establish the circumstances under which the nodes are used and may convey information as:

- the role of the node in the UNL graph, as for the attribute .*@entry*, that indicates the main (starting) node of a UNL graph

- the semantic and pragmatic information, which is conveyed in natural languages by various means such as flexion and derivation morphemes, or analytic constructions. For example, imperfect past in French expressing habit in the past can be noted as .@past.@repetition, and immediate future (I am coming! J'arrive!) as .@future.@immediate.

- the (external) context of the utterance, i.e., non-verbal elements of communication, such as speech-act, sentence and text structure, politeness level, schemes, etc.

UNL2010 specifications propose 344 universal attributes organized in a hierarchy. User can refer to the specifications at [14] or to RDF-UNL at [6] for the complete list.

**Implementation.** We define the datatype *unl:attribute* as the list of the 344 possible strings for universal attributes with the following code :

```
unl:attribute
  rdf:type rdfs:Datatype ;
  rdfs:comment """   More informations about Universal Attributes here
              http://www.unlweb.net/wiki/Universal_Attributes   """ ;
  rdfs:label "Universal Attribute" ;
  owl:equivalentClass [ rdf:type rdfs:Datatype ;
              owl:oneOf (  ".@1" ".@2" ".@3" ".@ability" ".@about" ".@above" […]
                        ".@worth" ".@yes" ".@zoomorphism")    ] .
```

As attributes are organized in a hierarchy, we defined subclasses of *unl:attribute* representing individual attribute values with their definitions. For instance:

```
<https://unl.tetras-libre.fr/rdf/schema#@hyperbole>
    rdf:type owl:Class ; rdfs:label "hyperbole" ; rdfs:subClassOf :Tropes ;
    skos:definition "Use of exaggerated terms for emphasis" .
```

At the moment those subclasses are informative only. They provide a way to include the attributes hierarchy and definitions but are not themselves declared as instances of *rdfs:Datatype* as we thought it would be an unnecessary complication.

Attributes are attached to UNL graph nodes using the datatype property *unl:has_unl_attribute* defined with the following code :

```
unl:has_attribute
    rdf:type owl:DatatypeProperty ; rdfs:domain :UNL_Node ; rdfs:range unl:attribute .
```

Further examples are shown in section 3, presenting how attributes are used to represent sentences as RDF-UNL graphs.

# 3     Representing sentences and documents as RDF-UNL

## 3.1    Scopes in UNL hypergraphs

As we already said, UNL may need more than simple graphs to represent a sentence. It may need hypergraphs where some nodes are another UNL graph, and not simply a UW. Such a hypergraph is presented in **Fig. 3**. In the sentence "The system displays a channel in green when it is in the broadcast state", when is linked by the relation obj to the entire sub-sentence "[a channel] is in the broadcast state", itself represented as a UNL subgraph. Those hypernodes are called scopes. The main scope of a graph is not explicitly declared. It is also important to note that a scope contains relations between UWs and not only UWs. Indeed, a UW occurrence can participate in several scopes, as illustrated in **Fig. 3**, where "channel" is in the main scope and scope 01. Each scope must contain exactly one UW bearing the attribute *.@entry*.

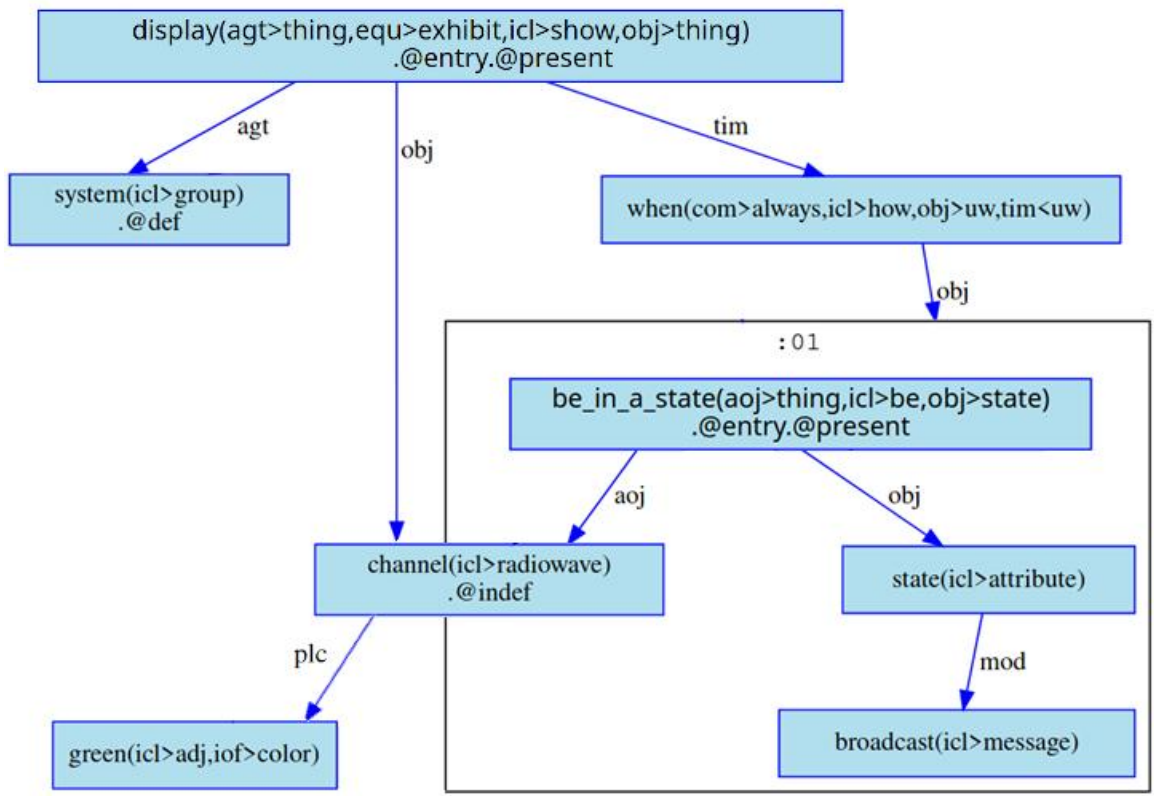

*Fig. 3. Example of a UNL hypergraph*

In the following, we discuss two implementations that have both advantages and drawbacks. The first one uses RDF *named graphs* and the second a *reification* of relations. At the moment we consider that the two possibilities can coexist in RDF-UNL, one or the other being used depending on specific application constraints. It is easy to write SPARQL rules to convert one implementation into the other.

**Implementation 1 – Named graphs.** An RDF graph can itself be named with an URI. However, the RDF standard does not provide a way to declare which triples belong to a graph, other than gathering them in the same file, accessible over HTTP at the graph's URI (in this case, a URL). Even though there are proposals to extend RDF triples as quads [15], adding the provenance graph to a triple, they are not part of the standard and their support in existing tools is very heterogeneous. Therefore, RDF-UNL allows representing scopes as named graphs, but also provides another means to handle them (reification), that we describe in the following paragraph. The following code presents a RDF-UNL graph, with scopes as named graphs, using TriG syntax [15] (triples that are in the scope are grayed):

```
example:R2
    rdf:type unl:UNL_Sentence ;
    rdfs:label "The system displays a channel in green when it is in broadcast state"@en .
example:UNL_Scope_00000017
    rdf:type unl:UNL_Scope ;
    rdfs:label "01"@fr ;
    unl:is_substructure_of :R2 .
example:when_00000012
    rdf:type unl:UW_Occurrence ;
    rdfs:label "when(icl>how,com>always,tim>uw,obj>uw)" ;
    unl:has_lexeme example:when-icl--how,com--always,tim--uw,obj--uw-
    unl:obj :UNL_Scope_00000017 .
example:UNL_Scope_00000017 {   example:be_in_a_state_00000013
                               unl:aoj exemple:channel_00000014 ;
                               unl:obj example:state_00000015 .
                               example:state_00000015
                               unl:mod example:broadcast_00000016 . }
```

**Implementation 2 – Reification.** Another way to declare that a relation belongs to a scope is to increase its arity. In this perspective, UNL relations link not only a source and a target, but also a scope. The only way to represent n-ary relations using RDF is using reification, as explained in [16].

To achieve this, we declared that the *unl:Universal_Relation* property and its sub-properties are also of type *owl:Class* and organized as sub-classes. To declare that a relation *ex:UW1 unl:rel ex:UW2* holds in scope *ex:scope1*, we write something like:

```
ex:rel1 a unl:rel ; unl :has_source ex:UW1 ; unl :has_target ex:UW2 ; unl :has_scope ex:scope1 .
```

In this setup, we not only declare that a relation holds, but we explicitly create an occurrence of this relation as an instance of *unl:Universal_Relation*. In the example we gave for implementation 1, the part *example:UNL_Scope_00000017{[...]}* becomes the following (the rest is not changed):

```
example:be_in_a_state_00000013--aoj--channel_00000014
    rdf:type unl:aoj ;
    unl:source example:be_in_a_state_00000013 ;
    unl:target exemple:channel_00000014 ;
    unl:has_scope example:UNL_Scope_00000017 .
example:be_in_a_state_00000013--obj--state_00000015
    rdf:type unl:obj ;
    unl:source example:be_in_a_state_00000013 ;
    unl:target exemple:state_00000015  ;
    unl:has_scope example:UNL_Scope_00000017 .
example: state_00000015--mod--broadcast_00000016
    rdf:type unl:mod ;
    unl:source exemple:state_00000015 ;
    unl:target example:broadcast_00000016 ;
    unl:has_scope example:UNL_Scope_00000017 .
```

## 3.2 Document structure

UNL allows a very simple structuring: documents are divided into paragraphs, themselves divided into sentences.

**Implementation:** We created the classes *unl:UNL_Document*, *unl:UNL_Paragraph* and *unl:UNL_Sentence*, which are structures larger than *unl:UNL_scope* and *unl:UW_Occurrence*. Instances of those five classes can be linked with the relations *unl:is_superstructure_of* and *unl:is_substructure_of*.

## 3.3 Existing UNL services

The task of creating UNL graphs from natural language texts is called *enconversion*. The reverse task is called *deconversion*. Such a bi-directional service, supporting English and Russian (usable for translation) is described in [17] and available online[8]. The UNL graphs of Fig. 1 and Fig. 3 are post-edited graphs obtained from this service. It is the easiest way to start playing with UNL.

UNL[Web][9] provides several online applications[9] and tools[10] around UNL, as well as linguistic resources for a dozen languages[11].

A tool to show UNL graphs in SVG and a "toy" French deconverter are also available online as part of Lingwarium[12]. Open source developments made for the UNseL project will also be released in Linguarium. This should include a French enconverter and deconverter.

Finally, as part of our present work, a UNL to RDF serializer (unl2rdf) has been developed and deployed as a Web service[13]. All the source code is available at [6].

---

[8] http://unl.ru/deco.html

[9] http://www.unlweb.net/wiki/Applications

[10] http://www.unlweb.net/wiki/Tools

[11] http://www.unlweb.net/unlarium/

[12] http://lingwarium.org, in the top left click Workplace, then select UNL, Language pair UNL-FRA and finally the test tab.

[13] https://unl.demo.tetras-libre.fr/

# 4 OWL axioms from RDF-UNL graphs and reasoning

## 4.1 Working example

In this section, we present a working example for a possible application of RDF-UNL that is currently experimented in the UNseL project. We consider the two system requirements of Fig. 1 and Fig. 3:

R1: The system allows a radio channel to take on two states: Listening and Traffic.

R2: The system displays a channel in green when it is in broadcast state.

The goal is to automatically detect the terminological incoherence between the two requirements: R1 declares that Listening and Traffic are the only allowed states for a radio channel whereas R2 talks about a channel in broadcast state.

To achieve this goal, we propose the following:

1. Enconvert R1 and R2 to obtain UNL graphs like Fig. 1 and Fig. 3 (they are post edited graphs generated by the Web service at http://unl.ru/deco.html),
2. Serialize the UNL graphs in RDF (using https://unl.demo.tetras-libre.fr/unl2rdf),
3. Use generic SHACL rules (W3C recommendation [18]) to construct OWL statements from the RDF-UNL graphs,
4. Use a generic OWL reasoner to detect inconsistencies.

## 4.2 Generic SHACL rules

We have developed and executed SHACL SPARQL rules (*sh:SPARQLrule*), using Topbraid Composer Free Edition. The rules are generic enough so they can be used in various contexts to construct OWL statement that can support reasoning.

**Rule 1: Extract cardinality.** This first rule apply on (reified) instances of the relation *unl:qua* used to express the quantity of an entity. It constructs a statement expressing an *owl:cardinality* with the following SPARQL code:

```
CONSTRUCT { ?lex owl:cardinality ?tInt }
WHERE { ?this   a unl:qua ;
               unl:has_source ?s ;
               unl:has_target ?t .
        ?t rdfs:label ?tLabel .
        ?s unl:is_occurrence_of ?lex
        BIND(xsd:integer(?tLabel) AS ?tInt)}
```

**Rule 2: Extract enumerations.** This second rule detect enumeration patterns in RDF-UNL graphs and declare the adequate class using *owl:oneOf*, supposing that the enumeration is complete. The SPARQL query also checks that the UNL restrictions of the "head"[14] of the enumeration contain *icl>attribute*. Hence we know that we are talking about possible values for an attribute and declare the enumerated class as an *owl:Datatype*. Finally, we did not find a query that works for arbitrary long enumerations. The following code works only for length two[15], as in R1:

```
CONSTRUCT {    ?lex  owl:equivalentClass [ rdf:type rdfs:Datatype ; owl:oneOf (?l1 ?l2)    ]    }
WHERE { ?this   a unl:cnt ;
           unl:has_source ?s ;
           unl:has_target ?t1 .
       ?s unl:is_occurrence_of ?lex .
       ?lex rdfs:label ?label .
       FILTER regex(?label,"icl>attribute")
       ?and1   a unl:and ;
           unl:has_source ?t1 ;
           unl:has_target ?t2 .
       ?t1 rdfs:label ?l1 .
       ?t2 rdfs:label ?l2 .    }
```

**Rules 3 and 4: Extract and instantiate datatype properties.** Finally, two rules identify patterns described in Fig. 4:

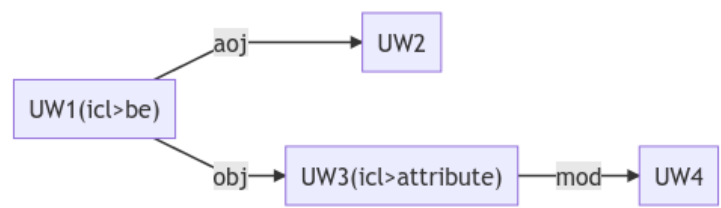

*Fig. 4.* A UNL pattern corresponding to an owl:DatatypeProperty

Rules 3 and 4 produce the following RDF/OWL triples (pseudo-code), where *UWi_lexeme* is the lexeme corresponding to *UWi* occurrence and UWi_label its label:

```
UW1_lexeme   a owl:DatatypeProperty ;
         rdfs:domain UW2_lexeme ;
         rdfs:range UW3 .
UW2_lexeme   a owl:Class .
UW3_lexeme   a rdfs:Datatype .
UW2          a  UW2_lexeme ;
         UW1_lexeme UW4_label .
```

---

[14] In R1, we call *state* the "head" of the enumeration *two states: Listening and Traffic.*

[15] We included code for longer enumerations in our public Git.

## 4.3 Reasoning on extracted facts

Altogether, the rules extract the following facts from the R1 and R2 graphs:

```
@prefix example: <https://unl.tetras-libre.fr/rdf/example#> .
example:be_in_a_state(aoj--thing,icl--be,obj--state)
    rdf:type owl:DatatypeProperty ;
    rdfs:domain example:channel(icl--radiowave) ;
    rdfs:range example:state(icl--attribute) .
example:channel(icl--radiowave)
    rdf:type owl:Class .
example:channel(icl--radiowave)__00000014
    rdf:type example:channel(icl--radiowave) ;
    example:be_in_a_state(aoj--thing,icl--be,obj--state) "broadcast(icl>message)" .
example:state(icl--attribute)
        rdf:type rdfs:Datatype ;
        owl:cardinality 2 ;
        owl:equivalentClass [
            rdf:type rdfs:Datatype ;
            owl:oneOf (
                "listening(icl>sensing)"
                "traffic(icl>communication)") ; ] .
```

The Ontology Debugger Plug-In for Protégé [19] executed on the previous triples finds a logical inconsistency (**Fig. 5**), that corresponds to the targeted incoherence. R1 declares that *Listening* and *Traffic* are the only allowed states for a radio channel, whereas R2 mention a channel being in *broadcast state*. To wrap things up informally, the following incompatible logical axioms have been extracted:

- the data type */state/* is the class *owl:oneOf* ("*listening*", "*traffic*"),
- */be_in_state/* is a property with range */state/*,
- the relation */channel/*--*/be_in_state/*→ "*broadcast*" also exists but is incompatible with the two previous axioms.

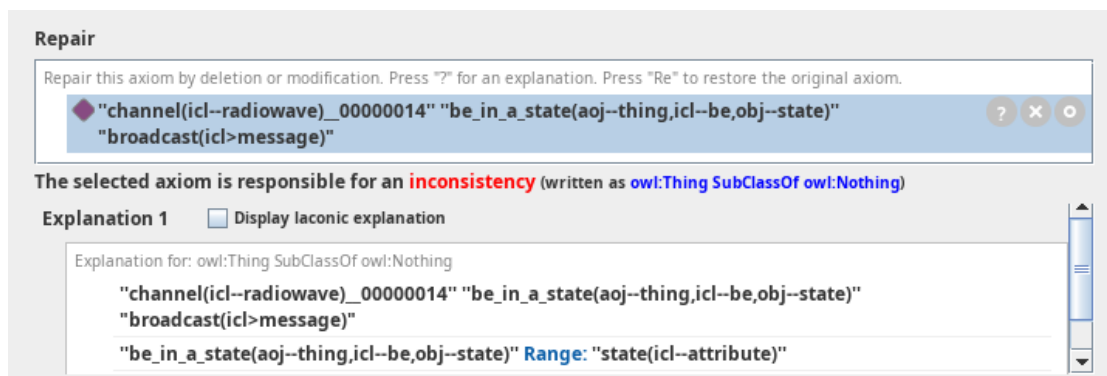

**Fig. 5.** *Result of the Ontology Debugger Plug-In for Protégé*

Page 15

## 5    Conclusion and future work

In this paper we presented RDF-UNL, an Open Source implementation of the Universal Networking Language on top of RDF. It equips the Semantic Web with a linguistic framework to soundly represent the meaning of a sentence, directly in RDF. The UNL is designed to handle multilingualism and proposes enconversion and deconversion services to go from natural languages to UNL graphs, and the other way around. Therefore, RDF-UNL is a very good middleware for any Semantic Web application that requires transformations between natural language and RDF (in one way, or the other), for instance: natural language querying of ontologies, description of axioms in natural language, content extraction from texts, etc.

We illustrated the usability of RDF-UNL to detect incoherence in system requirements. This preliminary example already shows that it is possible to construct meaningful OWL axioms that support non-trivial reasoning. In particular, using RDF-UNL as an intermediate step from texts to RDFS or OWL shall help us to overcome some limitations of existing systems: it becomes possible to handle semantic proximity of terms, to go beyond raw RDF by categorizing resources according to a schema (classes, instances, data types, etc.), and to reduce the monolingual dependency of Semantic Web software that must work in multilingual situations.

RDF-UNL and the application given as example in this article are currently being developed as part of the UNseL[16] project, funded by the French DGA (General Directorate of Defence). Future work will focus on (1) developing a high quality enconverter for French and English, dedicated to system requirement documents and (2) scaling up the construction of OWL axioms and reasoning using SHACL, as well as including domain or upper ontologies in the process.

---

[16]    UNseL : Universal Networking system engineering Language